\documentclass[11pt,letterpaper]{article}

\usepackage[margin=1in]{geometry}

\usepackage{amsmath,amssymb,amsfonts}
\usepackage{bm}
\usepackage{textcomp}
\usepackage[T1]{fontenc}

\usepackage{graphicx}
\usepackage{booktabs,multirow}
\usepackage[table]{xcolor}
\usepackage{multicol}

\usepackage{setspace}
\usepackage[numbers,sort&compress,super]{natbib}
\usepackage[hidelinks]{hyperref}

\newcommand{\acks}[1]{\section*{Acknowledgments}}
\newcommand{\ethics}[1]{}
\newcommand{\coi}[1]{\section*{Conflict of Interest}#1}
\newcommand{\data}[1]{\section*{Data Availability Statement}#1}

\begin{document}

\title{\bfseries Benchmarking transferability of SSL pretraining to same and different modality segmentation tasks}

\author{
Jue Jiang \quad Harini Veeraraghavan\\[0.4em]
\small Department of Medical Physics, Memorial Sloan Kettering Cancer Center\\
\small \texttt{jiangj1@mskcc.org} \quad \texttt{veerarah@mskcc.org}
}

\date{\small Preprint submitted to \textit{Medical Physics}}

\maketitle

\begin{abstract}
\noindent\textbf{Background:} Self-supervised learning (SSL) has emerged as an unsupervised pretraining approach that improves transferability and accuracy of transformer models for medical image analysis. A number of SSL pretext tasks have been proposed, but it remains unclear which method is most effective when applied to diverse medical imaging modalities and downstream segmentation tasks of varying complexity.

\vspace{0.5em}\noindent\textbf{Purpose:} To benchmark nine representative SSL methods pretrained on an identical, large-scale 3D CT dataset and evaluate their transferability for segmenting diverse anatomical structures and tumors from CT and MRI in both many- and few-shot fine-tuning settings.

\vspace{0.5em}\noindent\textbf{Methods:} Nine SSL methods spanning four pretext-task families---masked image modeling (SimMIM, SMIT), self-distillation (DINO, iBOT), contrastive learning, and rotation prediction---were pretrained from scratch using the same 10{,}412 3D CT scans (1.89~M 2D axial slices) covering varied disease sites. The pretrained Swin Transformer encoder from each method was integrated into a SwinUNETR-style segmentation network (Swin encoder with a 3D CNN decoder and skip connections) and fine-tuned on nine public segmentation tasks of varying complexity, including large abdominal organs and tumors from CT and MRI. Performance was assessed using Dice similarity coefficient (DSC). Transferability across modalities (CT-to-MRI), and feature-reuse patterns between few- and many-shot fine tuning were further analyzed using centered kernel alignment.

\vspace{0.5em}\noindent\textbf{Results:} Self-distilled masked image transformer (SMIT), which combines masked image modeling (MIM) with local and global self-distillation, achieved the highest overall segmentation accuracy across the nine tasks, and the smallest few-shot-to-many-shot performance gap, indicating the strongest data efficiency. SMIT also showed the most consistent feature-reuse patterns between few- and many-shot fine tuning. MIM-based SimMIM and self-distillation methods (DINO, iBOT) outperformed contrastive learning and rotation prediction, which rely on image-level global representations. Differences between SSL methods were largest in the few-shot setting and narrowed as the size of the labeled fine-tuning dataset increased, indicating that the choice of SSL pretraining matters most under limited annotation budgets.

\vspace{0.5em}\noindent\textbf{Conclusions:} Among the nine SSL methods benchmarked under matched pretraining data and architecture, SMIT provided the best transferability for medical image segmentation, including across modalities. Dense pretext tasks that combine local masked prediction with self-distillation are more effective for downstream dense prediction than global image-level objectives. Performance differences between SSL methods diminish with larger labeled datasets. Code and pretrained/fine-tuned model checkpoints will be released through GitHub upon manuscript acceptance.

\vspace{1em}\noindent\textbf{Keywords:} Self-supervised learning, benchmarking, pretrained medical segmentation models, transformers.
\end{abstract}


\section{Introduction}
Self-supervised learning (SSL) is an unsupervised pretraining method that self-labels images to learn easily transferable feature representations for various downstream tasks. SSL models followed by supervised fine-tuning have been shown to often outperform training from scratch for medical image analysis~\citep{Willemink2022,Haghighi2024_MedIA}, although the magnitude of improvement varies with the pretraining task and the amount of downstream labeled data.

SSL extracts useful feature representations by creating ''pretext tasks" that use images themselves as labels. In several settings, SSL has been reported to be more accurate than supervised pretraining, in part because SSL methods extract richer and more diverse feature embeddings that can be repurposed to various imaging modality tasks~\citep{pmlr-v158-truong21a, Ericcson2021,Reed_2022_WACV, gomez2024swintransformers, Goyal2021}. However, increasing difference of the target domain from the pretrained data domain degrades accuracy~\citep{Goyal2021, Matsoukas2022,Reed_2022_WACV}. Staged pretraining that refines generalist models pretrained with photographic images using domain- or task-specific datasets can mitigate domain differences~\citep{kalapos2023,Reed_2022_WACV}. Another common approach creates task-specific pretrained models using unlabeled task-specific datasets followed by supervised fine tuning~\citep{DufumierMICCAI2021,ZHANGMedIA_2023,zhou2021models,taleb20203d}. Recent works employ a computationally efficient pretraining once with numerous radiological imaging datasets sourced from multiple disease sites, which then can be finetuned to both CT and MR-based segmentation tasks~\citep{hatamizadeh2022_brain,jiang2022self,Willemink2022,jiang2025_MedPhys, Paudyal2024}. However, little is understood about how the choice of SSL tasks impacts transferability to different modalities and when SSL pretraining is beneficial for downstream tasks. 

Hence, the goal of this work is to benchmark various SSL pretraining methods applied to 9 different segmentation tasks of varying complexity ranging from multi-organ to tumors segmentation on same and different modality than pretraining. First, all SSL methods were created using an identical pretraining dataset of 10,412 3D CTs spanning anatomical sites from head to the pelvis involving various diseases, followed by fine tuning on identical public datasets to evaluate methods under same conditions. In this respect, the significance and impact of our work lies in providing a suite of pretrained SSL architectures using identically large pretraining datasets for further implementations on different disease sites by others. Second, extending prior works that applied pretrained models to other modalities in many-shot ($>$20 and $\leq$ 100 cases) settings~\citep{hatamizadeh2022_brain,Haghighi2024_MedIA,Paudyal2024}, the current work explicitly studies how SSL impacts transferability, feature reuse, training efficiency, and accuracy in many- and few-shot settings in different modality than pretraining. Instead of developing new architectures and learning tasks, the novelty of our approach lies in deriving insights and understanding under what conditions SSL pretraining is beneficial, what SSL tasks have universal downstream applicability, and why pretraining is effective for different downstream tasks. 
The current work is a significant extension of our prior works that (a) introduced SMIT pretraining for multi-organ segmentation with a Swin Transformer encoder~\citep{jiang2022self}  and (b) recent work that studied feature reuse and transferability of a subset of 4 SSL methods using ViT for segmenting abdominal organs from CT and MRI~\citep{Jiang_ISBI_2025}. The improvement and our contributions include:
\begin{itemize}
    \item Benchmarking analysis to study impact of multiple SSL methods pretrained using identical and numerous, loosely curated 3D CT dataset of 1.89M 2D axial slices extracted from 10,412 3D CTs arising from various cancer and non-cancers spanning head and neck, chest, and abdomen, providing a basis to evaluate models without influence of pretraining data variations. All models and fine tuned checkpoints will be provided through GitHub upon manuscript acceptance.
    \item Comprehensive analysis was performed to identify empirically effective SSL method applicable to segmentation tasks of varying complexities involving 32 structures including organs and tumors from two different modalities (CT and MRI). Performance differences due to pretraining data, model, few- and many-shot fine tuning data were also evaluated.  
    \item Finally, we studied how pretext tasks impact transferability of the same set of tasks including multi-organ and tumor segmentation across modalities by analyzing feature reuse under many- and few-shot fine tuning settings.
\end{itemize}

    	\begin{figure*}[!htb]
		\begin{center}
			\includegraphics[width=\textwidth]{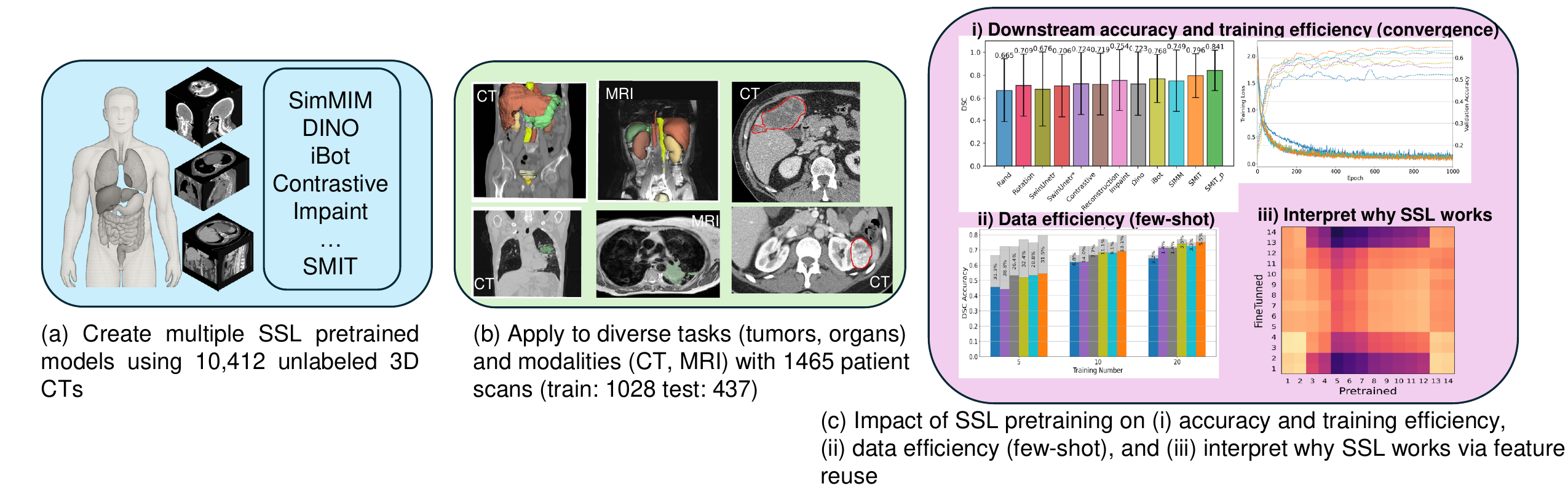}
			\caption{\small (a) and (b) illustrate the SSL pretraining methods and downstream tasks. (c) summarizes the analyses conducted in this paper.} \label{fig:overview}
		\end{center}
	\end{figure*}

\section{Related works}
SSL pretraining is a highly effective method for medical image analysis tasks including segmentation~\citep{jiang2022self,jiang2024selfsupervised,Haghighi2024_MedIA,zhou2021models,Willemink2022,kalapos2023}, detection and classification~\citep{Haghighi2024_MedIA,ZHANGMedIA_2023,Goyal2021,Matsoukas2022}. Detailed overview of SSL methods for medical image analysis are covered in prior works~\citep{ZHANGMedIA_2023,manna2024selfsupervised}. We discuss works pertaining to our key focus related to SSL for segmentation and transferability across modalities.  

\subsection{SSL pretext tasks for medical image analysis}

Contrastive pretraining is a commonly used approach due to its robustness to occlusions and image rotations~\citep{sriram2021covid,chaitanya2020contrastive,zhu2020embedding,xie2020pgl,caron2021emerging}. It 
extracts global image context by learning the similarity of augmented image views created from an input image (positive pairs) while increasing distance between views created from different examples~\citep{he2020momentum}.  However, contrastive pretraining may fail to learn useful representations if images are not "object-centric", wherein each example contains only one class. Medical images contain distinctly different anatomic regions, exacerbating the difficulty to separate individual anatomic regions when learning with cropped views~\citep{Hassanpour2024OvercomingDC}. Ultimately, accuracy of contrastive pretraining for  segmentation is limited~\citep{chaitanya2020contrastive}. 

Image reconstruction tasks recover an original image from its corrupted version by extracting the dense local spatial and anatomic context, which benefits segmentation. Representative image reconstruction pretext tasks include jigsaw puzzles~\citep{zhu2020rubik,taleb20203d}, masked image slice prediction~\citep{jun2021medical}, masked patch reconstruction~\citep{zhou2021models}, context restoration~\citep{chen2019self,feng2020parts2whole,zhou2023unified_PRCLV2}, and relative image position prediction~\citep{komodakis2018unsupervised}. Semantic genesis~\citep{haghighi2020learning} improved over reconstruction methods by including a self-learned anatomic prior via a self-classification network that extracted similarities of individual anatomic regions between different patients. Masked image modeling (MIM) is a reconstruction approach that was designed for transformers wherein uncorrupted images are extracted from a randomly masked sequence of image patch tokens~\citep{li2021mst,xie2021simmim,zhou2022image,chen2023masked, bao2022beit,zhou2022image,he2021masked}. 

Unlike knowledge distillation that seeks to transfer knowledge contained in a high capacity ''teacher'' network to a lower capacity ''student'' network, self-distillation learning seeks to smooth extracted feature variations and improve the relevance of extracted feature embeddings by adding a regularization provided by an exponentially moving averaged (EMA) teacher network~\citep{xue2023stare}. Recent works have shown improved accuracy with self-distillation pretraining~\citep{caron2021emerging, zhou2022image,li2021mst}. 

Recent works combined multiple pretext tasks to extract global and local feature representations~\citep{zhou2023unified_PRCLV2,jiang2022self,tang2022self,Haghighi2024_MedIA,taleb20203d}. Importantly, combining diverse pretext tasks has shown to improve robustness in small-data regimes and feature reuse especially in low and mid-level layers~\citep{Haghighi2024_MedIA,jiang2022self,jiang2024selfsupervised}. However, the interplay between the pretext tasks and architecture to effectively leverage pretraining data for learning features that are transferable not just to similar domains (CT to CT) but to different domains (CT to MRI) is not well studied, one key focus of this work. 

\subsection{Transferability of SSL pretraining to downstream tasks}

Success of both supervised~\citep{Huix2024, Goyal2021, Matsoukas2022} and SSL pretraining~\citep{jiang2022self,Haghighi2024_MedIA,ZHANG2024102996} to downstream tasks has been linked to feature transferability and reuse particularly in low and mid-level layers. It is also known that accuracy degrades as the distance of the downstream domain (e.g. X-ray images) increases with respect to the pretraining domain (e.g. photographic images)~\citep{Matsoukas2022,Ericcson2021}. Also supervised pretraining has shown to overfit features to upstream tasks that reduce intra-class diversity~\citep{feng2022rethinking} and richness of feature embeddings compared to SSL methods~\citep{pmlr-v158-truong21a}, resulting in lower accuracy. On the other hand, SSL pretraining enhances models' effectiveness to extract object boundaries and contextual information~\citep{caron2021emerging}, which increases accuracy~\citep{haghighi2020learning} and robustness to varied image acquisitions~\citep{jiang2025_MedPhys}. 
\\
However, the impact of modality differences for the same tasks on feature reuse from choice of SSL tasks is unstudied. The current work evaluated the variations in patterns of feature reuse between modalities for distinctly different tasks such as organs and tumor segmentation from CT and MRI under few- and many-shot fine tuning as well as diversity of attention to spatial context such as organs and the resultant  impact on accuracy.

\subsection{Benchmarking SSL impact for medical image tasks}
Prior works benchmarked effectiveness of generalist SSL models for varied tasks including medical images, and showed that increasing distance of target domain from pretraining degraded accuracy~\citep{Ericcson2021,Goyal2021}. Recent studies have shown that medical image pretrained models are accurate than generalist models for medical image segmentation tasks~\citep{wen2021rethinking}. Importantly, increasing the size of medical datasets for pretraining is crucial to enhance accuracy compared to generalist models especially in low data regime tasks~\citep{haghighi2021transferable,xie2022unimiss}. 

Hence, recent works focused on creating large scale medical image pretraining datasets~\citep{suprem2024,Butoi2023,Ma2024}, demonstrating accuracy improvements over generalist pretraining and scratch training. Representative works employing large scale SSL pretraining with medical images include masked encoder  approach using$\sim$39k MRI volumes ~\citep{wald2025revisiting}, SwinUNETR~\citep{hatamizadeh2021unetr}, as well as our prior work~\citep{jiang2025_MedPhys,Jiang_ISBI_2025}. Whereas these methods provided strong evidence that large-scale, single modality-based pretraining can enhance accuracy, none to our best knowledge have studied how downstream training efficiency, model transferability and feature reuse to a different modality is impacted by SSL.

\section{Method}
\subsection{Goal} To benchmark and identify an empirically accurate SSL approach for medical image segmentation tasks by creating a suite of pretrained transformer encoders using various SSL methods from identical, large and unlabeled CT datasets involving diseases from head to pelvis. Figure~\ref{fig:overview} depicts an overview of our benchmarking approach. 

\subsection{Approach}
Nine different SSL pretext tasks, shown to be useful for segmentation were evaluated. The evaluated SSL tasks are: 
\begin{enumerate}
\item SimMIM~\citep{xie2021simmim}: The basis of SimMIM is MIM, which requires the model to reconstruct raw pixel values underlying randomly masked patches (typically 75\% of image). SimMIM has shown to be highly effective for extracting robust visual representations and achieving competitive performance in several vision tasks. A lightweight decoder was connected to the transformer encoder to perform dense pixel intensity regression. L2 norm was used to optimize the network as: 

\[
\mathcal{L}_{\text{SimMIM}} = \frac{1}{|M|} \sum_{i \in M} \left\| x_i - \hat{x}_i \right\|_2^2,
\]
where \( M \) denotes the set of masked patches, \( x_i \) is the original voxel value, and \( \hat{x}_i \) is the predicted value.
 
\item Inpainting~\citep{tang2022self}: Inpainting reconstructs whole images from randomly masked regions in input images. Hence, unlike SimMIM, accuracy of both visible and masked image regions are computed, in order to capture both the global structure and local semantics of the image. This loss is computed as:
    \[
\mathcal{L}_{\text{Inpaint}} = \frac{1}{N} \sum_{i=1}^{N} \left\| x_i - \hat{x}_i \right\|_2^2,
\]
where \( N \) is the total number of voxels in the image.
\item Reconstruction: Reconstruction generates images through dense pixel regression from original uncorrupted images. Hence, the goal is to learn a robust latent representation of the images by capturing the low-level textures and global structural information. This task is optimized as:
\[
\mathcal{L}_{\text{recon}} = \frac{1}{N} \sum_{i=1}^{N} \left\| x_i - \hat{x}_i \right\|^2.
\]
\item Contrastive Learning~\citep{tang2022self}: Different from generative methods like SimMIM, a discriminative learning framework extracts feature representations that maximize the distance between positive (or augmented views of same image) and negative samples (or different images) using infoNCE loss~\citep{oord2018representation} as:

   \[
\mathcal{L}_{\text{contrast}} = -\log \frac{\exp\left(\text{sim}(v_i, v_j)/t\right)}{\sum_{k=1}^{2N} \mathbf{1}_{[k \ne i]} \exp\left(\text{sim}(v_i, v_k)/t\right)},
\]

where \( v_i \) and \( v_j \) are embeddings of a positive pair, \( \text{sim}(\cdot,\cdot) \) denotes cosine similarity, \( t \) is a temperature parameter, and \( \mathbf{1}_{[k \ne i]} \) is an indicator function that excludes the anchor from the denominator. \( 2N \) denotes the number of augmented views in a batch, where each sample has two views.

\item Rotation~\citep{tang2022self}: The goal of this task is to predict one of predefined rotations (e.g., 0°, 90°, 180°, 270°) applied to the input 3D image in order to extract orientation-specific features. A linear probing layer was implemented following the transformer encoder to output the predicted rotation \( \hat{y}_c \). Multi-category cross entropy loss for \( C \) rotation classes was used to optimize the network:
\[
\mathcal{L}_{\text{rotation}} = - \sum_{c=1}^{C} y_c \log(\hat{y}_c).
\]

\item DINO~\citep{caron2021emerging}: 
DINO employs a teacher-student self-distillation framework without requiring labels. The EMA teacher \( \theta_s \) is created from the student \( \theta_t \). The training objective encourages the student to extract similar token representations for the whole image as the teacher by minimizing cross-entropy loss between their softmax distributions as:

\[
\mathcal{L}_{\text{DINO}} = - \sum_{i=1}^{N} P_t(v_i, \theta_t) \log \left( P_s(\tilde{v}_i, \theta_s) \right),
\]

where, \( v_i \) and \( \tilde{v}_i \) denote two different views of the same image sample, \( P \) represents the probability distribution, and \( N \) is the number of image samples or views. 

\item iBOT~\citep{zhou2022image}: Image Bert pretraining with online tokenizer (BERT) combines MIM with a co-distillation framework similar to DINO. The MIM task is akin to explicitly introducing noise into the image patch tokens, thereby making the self-distillation task more challenging. The dual objective used in iBOT encourages the model to extract a spatially rich representation encompassing global image-level and local contextual features as:
\[
\mathcal{L}_{\text{iBOT}} = \lambda_g \mathcal{L}_{\text{global}} + \lambda_p \mathcal{L}_{\text{patch}}
\]

Global loss is computed as:
\[
\mathcal{L}_{\text{global}} = - \sum_{i=1}^{N} P_t(v_i, \theta_t) \log \left( P_s(\tilde{v}_i, \theta_s) \right).
\]

The patch-level loss learns local feature representations by encouraging the student network \( \theta_s \) to predict patch-wise probability distributions \( P_s^{\text{Patch}} \) using masked patches \( \hat{u}_i \) and aligned to \( P_t^{\text{Patch}} \) extracted by a EMA teacher \( \theta_t \) processing unmasked input views \( u_i \) as:

\[
\mathcal{L}_{\text{patch}} = - \sum_{i=1}^{N} m_i \cdot P_t^{\text{Patch}}(u_i, \theta_t) \log \left( P_s^{\text{Patch}}(\tilde{u}_i, \theta_s) \right),
\]

where, \( m_i \in \{0,1\} \) is a binary mask indicating whether patch \( i \) is masked, and \( N \) is the total number of patches.

\begin{table*}[!ht]
    \centering
\scriptsize
\caption{Fine-tuning datasets used for various segmentation tasks with training (full-shot) and test sets indicated.}
\label{tab:fine_tuning_data}
    \begin{tabular}{ccccc}
         Task &  Modality &  Disease Site & Training & Testing \\ \hline
         Organs&  CT&   Abdomen (AMOS) \citep{ji2022amos} &  100  & 100 \\
         Organs&  T2W MRI&   Abdomen (AMOS) \citep{ji2022amos}& 40 & 20 \\
         Tumor & CT  & Lung (TCIA)\citep{aerts2015data} & 377  & 196 (Internal)\\
         Tumor & T2wTSE MRI  & Lung (Internal) & 64 & 17 \\
         Tumor & CT & Liver (LiTS) \citep{bilic2023liver} & 101 & 30 \\
         Tumor & CT & Kidney (KiTS) \citep{heller2023kits21} & 400 & 91 \\
        \hline
        Total & CT/MRI & Abdomen, lung & 1,028 & 437 \\
        \hline
    \end{tabular}
\end{table*}
\item SMIT~\citep{jiang2022self}: Self-distilled masked image transformer (SMIT) employs multi-objective pretraining that combines MIM with token self-distillation tasks using a co-distilled EMA teacher network. This method combines three different pretext tasks that include masked image prediction (MIP), local or masked patch token distillation (MPD) and global image token distillation (GTD). The MIP task is similar to SimMIM and generates images underlying the randomly masked image tokens by using a linear projection layer connected to the transformer encoder using dense pixel regression. Unlike SimMIM, SMIT includes MPD and GTD losses that encourages the student network to extract robust feature representations aligned with features extracted by the teacher network. The key difference between MPD and GTD is that the former measures the feature token dissimilarity in the masked tokens while the latter focuses on the extracted global representation. Similar to DINO, temperature scaling is performed for both MPD and GTD tasks in order to soften the token distribution of the teacher network and improve effectiveness of distillation learning. In addition, a larger Swin model ([2,2,40,4] layers) pretrained using SMIT and called SMIT Plus (or SMIT$_{\mathrm{p}}$). SMIT$_{\mathrm{p}}$ that accepts inputs of size 128$\times$128$\times$128 voxels was evaluated to assess benefit of a bigger model against standard SMIT model.

\item SwinUNETR~\citep{tang2022self}: SwinUNETR uses multi-objective pretraining combining rotation prediction, reconstruction-based inpainting, and contrastive learning to extract both local texture details and global semantic representations from unlabeled 3D CT volumes. The original SwinUNETR was pretrained on 5,050 CT volumes. In addition to the publicly provided checkpoint an additional model was pretrained using the 10K dataset as used for all other models, and referred to as SwinUNETR$^*$.
    
\end{enumerate}

\subsection{Implementation details}
\subsubsection{Pretraining} 
All networks were implemented using the Pytorch library and trained on 4 $\times$ Nvidia GTX A100 with 4 $\times$ 80GB memory. Differing from the published SMIT model, which used a patch size of 128$\times$128$\times$128 voxels~\citep{Jiang_ISBI_2025,jiang2025_MedPhys}, all models in this work used a smaller patch size of 96$\times$96$\times$96 voxels for equitable comparison to the published SwinUNETR checkpoint~\citep{hatamizadeh2021unetr}. Models were optimized using ADAMw, cosine learning rate scheduler, trained for 500 epochs with an initial learning rate of 0.0002 and warmup for 30 epochs. Hyperparameters $\tau_{s}$ and $\tau_t$ were set to 0.1 and 0.07. $\tau_t$ was linearly warmed up from 0.04 to 0.07 in the first 30 epochs. GPU limitation was addressed for training, fine-tuning, and testing by resampling the original images to a voxel size of 1.5 mm$\times$1.5 mm $\times$2 mm. 
\\
The 3D Swin Transformer encoder~\citep{liu2021swin}, used as the backbone within a SwinUNETR-style segmentation network (encoder with 3D CNN decoder and skip connections) for all evaluated SSL methods, employed a hierarchical encoder of depth [2,2,8,2] across four stages with [4,4,8,16] attention heads. Input volumes were processed with a patch size of $2 \times 2 \times 2$ voxels, a window size of $4 \times 4 \times 4$ voxels, and a feature embedding size of 384. 
\\
For pretraining, all input CT scans were clipped within a predefined range [-500, 500] HU and then normalized to [0,1]. Augmented views for iBOT, DINO, SMIT, as well as contrastive pretraining was produced by randomly cropping 96$\times$96$\times$96 pixels from whole images. A default mask ratio of 0.75 was used for SMIT, SimMIM, and iBOT. Centering and sharpening operations reduced chances of degenerate solutions for token self-distillation tasks~\citep{caron2021emerging}. SMIT model hyperparameters including $\lambda_{MPD}$=0.1, $\lambda_{ITD}$ =0.1 were used as in the prior works to avoid refinement for evaluated tasks in this work. 

\subsubsection{Finetuning}
Segmentation networks were created by combining pretrained Swin Transformer encoder blocks with 3D convolutional decoders and skip connections, following the SwinUNETR design. Encoder features at each stage were concatenated with the corresponding decoder layers, which consisted of convolution, batch normalization, and LeakyReLU operations. An additional convolutional layer was employed to bridge the encoder features into the first decoder layer. The resulting Swin-encoder segmentor used a total of 64.70~M parameters. 
\\
Training was performed by using cropped regions of size 96$\times$96$\times$96 voxels containing various organs or tumors of interest. Segmentation for the entire image volume during testing was obtained using a sliding window strategy with 0.5 window overlap~\citep{Xie_CoTr_MICCAI21,hatamizadeh2021unetr}. Models were created to measure accuracy in full-, data-limited ($>$20 and $\leq$ 100), and few-shot training (5, 10, 20 examples) regimes. 
\\
Models were optimized using the ADAMw optimizer with a cosine learning rate schedule, starting at an initial learning rate of $2 \times 10^{-4}$ using a batch size of 3 to balance GPU memory constraints and training efficiency. Early stopping was applied to mitigate overfitting. Balanced class distribution for tumors were ensured by sampling 96$\times$96$\times$96 voxel cropped regions containing tumor or no tumor in a 1:1 ratio.  
\\
The CT images were normalized to [0,1] following intensity clipping within soft-tissue window of [-175 HU to 250 HU]. MR images for organ segmentation were normalized [0,1] following intensity clipping using 5\% and 95\% signal intensity values. MRIs were intensity normalized [0,1] following intensity clipping [0, 400] for lung tumor segmentation. 

\subsection{Datasets}
A total of 11,861 3D CT/MR volumes were analyzed. Models were pretrained using unlabeled datasets that did not overlap with the fine-tuning datasets. Separation of pretraining from fine-tuning datasets was intentional in order to assess the effect of pretraining using datasets completely unrelated to downstream tasks on feature reuse.   

\subsubsection{Pretraining}
Pretraining used 10,412 unlabeled 3D CT scans sourced from public~\citep{xiao2023lesion,harmon2020artificial,roth2015deeporgan}  and institutional cases as used in our prior work ~\citep{jiang2025_MedPhys,Jiang_ISBI_2025}. The scans were acquired with and without contrast, different vendors, and institutions, and for cancer and non-cancer diseases occurring from head to the pelvis. The data was used as is with no additional curation. No MRI scans were used for pretraining, which allowed evaluation of SSL transferability to a different modality. 

\subsubsection{Finetuning for downstream tasks}
The datasets used in the various downstream tasks are summarized in Table~\ref{tab:fine_tuning_data}. All tasks except lung tumor segmentation from MRI used public datasets to evaluate models' performance in datasets subject to distribution shift due to different scanner acquisitions. An internal CT dataset of patients with lung cancers undergoing immunotherapy was additionally used to assess performance in different cohort of patients than used in training (pretreatment CT of lung cancer patients treated with radiotherapy from the MAASTRO clinic, the Netherlands). 
\\
The primary datasets consisted of the AMOS CT and MRI~\citep{ji2022amos} and lung tumor CT and MRI datasets for assessing segmentation accuracy in few- and many-shot settings, domain transferability, and assessing patterns of feature reuse across the modalities. Secondary datasets included publicly available kidney (KiTS)~\citep{heller2023kits21} and liver cancer (LiTS)~\citep{bilic2023liver}, for evaluating few- and many-shot performance using the best performing methods identified in the primary datasets. Design experiments were performed using the primary datasets and best performing methods. Multi-organ segmentation models were optimized using a mixture of multi-category cross-entropy and DSC loss while tumors segmentation was optimized using binary cross-entropy and DSC loss.
\\
Analyzed MRI sequences included T2-weighted MRI used for abdominal organs and T2-weighted turbo spin echo (T2wTSE) for lung cancers. Lung cancers included primary and intra-thoracic metastatic lesions, ranging from 1.88 cc to 1,033 cc (training) and 2.96 cc to 413 cc (testing) in size. Liver cancers included primary (e.g., hepatocellular carcinoma, cholangiocarcinoma) and metastatic 
lesions acquired prior to and following any treatment, and were acquired from 7 different institutions. Hence, CTs varied widely in resolution, slice thickness (0.45 mm to 6.0 mm), and tumor burden (0–12 tumors per scan), reflecting real-world diversity in image acquisition and disease presentation. The kidney cancer dataset was collected in a single institution with clinical delineations performed on contrast-enhanced corticomedullary-phase CTs. 

\section{Experiments}
\subsection{Multi-organ segmentation}
Transferability of pretrained models to CT and MRI domains were measured by training separate model instances in a data-limited regime containing the same number (n = 40) of CT and MRI cases. Organs were grouped to assess accuracy of easy to segment large abdominal organs (liver, spleen, left kidney, right kidney), challenging to segment digestive organs that depict large appearance and anatomic configuration changes (esophagus, stomach, pancreas, duodenum, and gall bladder), small organs (right and left adrenal glands), vessels (aorta, inferior vena cava), and pelvic organs (bladder, prostate). Fine-tuning was performed in 1000 and 3000 epochs for full-/many-shot and few-shot regimes, respectively.  

\subsection{Tumors segmentation}
Separate models were trained for full (Table~\ref{tab:fine_tuning_data}) and few-shot (5, 10, 20) regimes for segmenting lung, liver, and kidney tumors and CT as well as MRI (for lung tumors). In addition, many-shot training was performed on pretrained models with 100 CT and 64 MRIs to assess transferability of lung cancer segmentation across the two modalities with similar number of cases. Full/many-shot training was conducted for 1,000 epochs. Few-shot training was performed for 3,000 epochs. Early stopping was applied to prevent overfitting.

\subsection{Metrics}
Segmentation accuracy was measured Dice similarity coefficient (DSC). Organ- and lung tumor-specific performance gap between CT and MRI modalities were measured by computing the difference in average DSC achieved on CT and MRI. Differences in DSC accuracies due to pretraining strategy was measured using two-sided, paired Wilcoxon signed rank test at 95\% confidence level for the various downstream tasks.

\subsubsection{Feature similarity analysis to assess transferability}
Transferability of pretrained to finetuned features were evaluated by computing the similarity of extracted features using centered kernel alignment (CKA). CKA computes a normalized similarity of two feature representations $\bm{X}$ and $\bm{Y}$ in terms of the Hilbert-Schmidt Independence Criterion (HSIC):
\begin{equation}
\setlength{\abovedisplayskip}{1pt}
\setlength{\belowdisplayskip}{1pt} 
 \mathrm{CKA}(\bm{K},\bm{L}) = \frac{\mathrm{HSIC_0}(\bm{K},\bm{L})}{\sqrt{\mathrm{HSIC_0}(\bm{K},\bm{K})\mathrm{HSIC_0}(\bm{L},\bm{L})}}
\label{eqn:CKA}
\end{equation}  
where $\bm{K}$=$\bm{X} \bm{X^T}$ and $\bm{L}$=$\bm{Y} \bm{Y^T}$ are the Gram matrices of feature $\bm{X}$ and $\bm{Y}$. 
In order to account for memory intensive calculations, a minibatch CKA~\citep{nguyen2020wide} performed by averaging HSIC scores over k minibatches was implemented followed by unbiased estimation~\citep{song2012feature} to reduce dependency of CKA values on the batch size.

\section{Results}

\begin{figure}[tp]
		\begin{center}
			\includegraphics[width=0.85\textwidth]{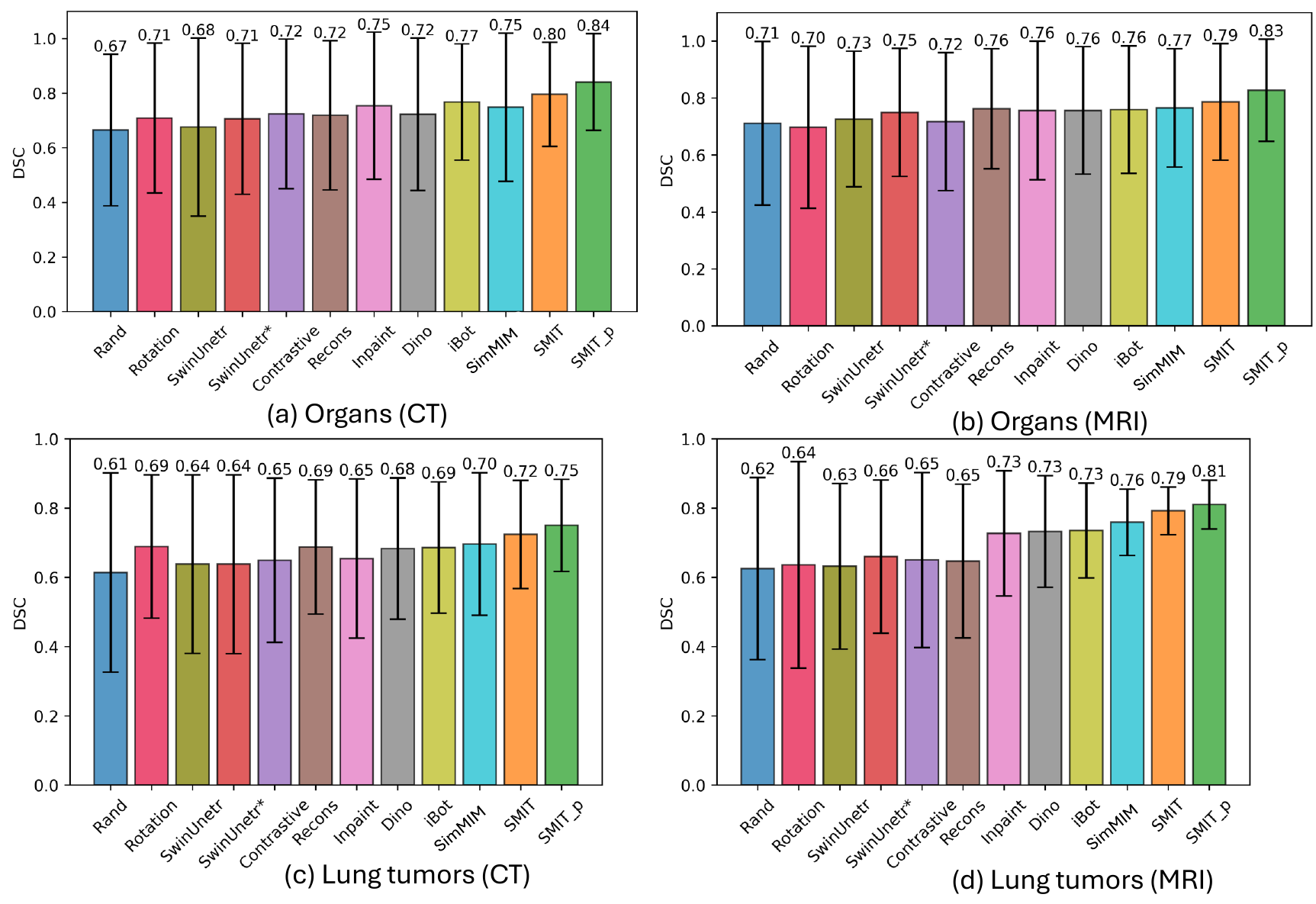}
			\caption{\small Impact of SSL pretraining methods applied to downstream segmentation tasks involving CT and MRI using the SwinUNETR-style segmentation network (Swin Transformer encoder with CNN decoder).} \label{fig:over_all_acc}
		\end{center}
	\end{figure} 

\subsection{Multi-organs segmentation}
 Figure~\ref{fig:over_all_acc} (a,b) shows the multi-organ segmentation accuracies from CT and MRI produced by the various models. Accuracies were averaged across all 15 organs for CT and 13 for MRI for simplicity of presentation. SSL methods using MIM, including SMIT and SimMIM outperformed all other methods in both modalities. SMIT was the most accurate with an average accuracy of 0.80 for CT and 0.79 for MRI. Self-distillation based SSL methods such as the iBOT and DINO were less accurate than SMIT and SimMIM, but more accurate than methods learning a global representation such as contrastive, reconstruction, and rotation prediction. Inpainting was more accurate than methods learning global representation for CT but similarly accurate for MRI. Although SwinUNETR$^*$ also employs a multi-objective pretraining, this method was less accurate than MIM-based and self-distillation methods for CT and MRI tasks.

\subsection{Lung tumor segmentation}
The SMIT model outperformed all other methods for segmenting lung tumors from CT (Figure~\ref{fig:over_all_acc}(c)) and MRI (Figure~\ref{fig:over_all_acc} (d)), with a DSC of 0.72 and 0.79, respectively. SimMIM was the second most accurate method after SMIT for both modalities. Token self-distillation methods such as iBOT and DINO were more accurate than contrastive, rotation prediction, and reconstruction methods for MRI; accuracy was similar for aforementioned methods on CT. Inpainting was less accurate than reconstruction pretraining for CT but was more accurate compared to the same method for MRI. SwinUNETR$^*$ was less accurate than MIM and self-distillation methods in both CT and MRI. 


\begin{figure}[!htp]
		\begin{center}
    \includegraphics[width=0.85\textwidth]{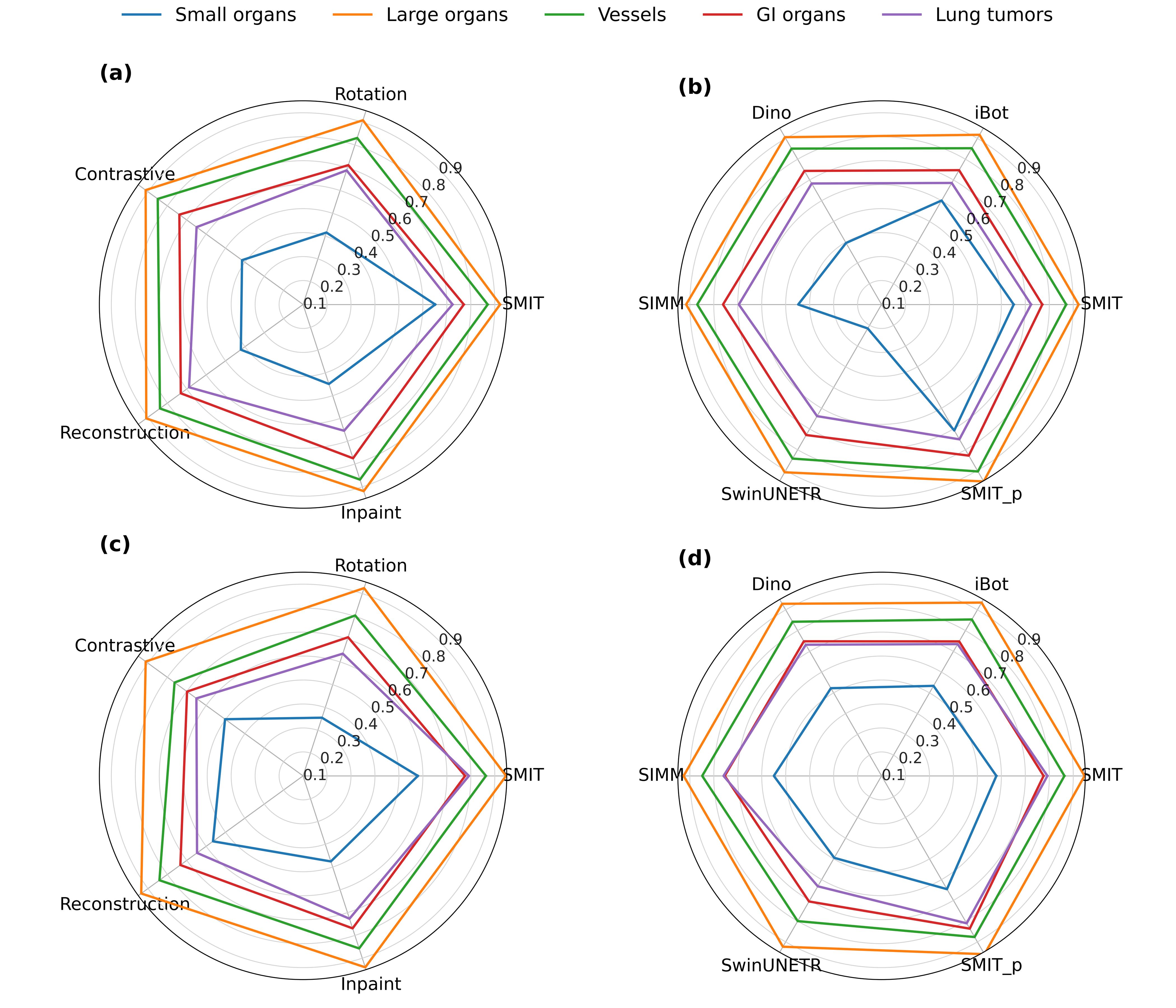}
			\caption{\small Grouped accuracies for small (Left adrenal, right adrenal, gall bladder), large (liver, left kidney, right kidney, spleen), gastrointestinal (GI) organs (stomach, duodenum, pancreas, esophagus), and lung tumor using SSL methods for CT (a,b) and MRI (c,d).} \label{fig:radar_plot}
		\end{center}
	\end{figure}
    
\begin{figure*}[!htp]
		\begin{center}
\includegraphics[width=\textwidth]{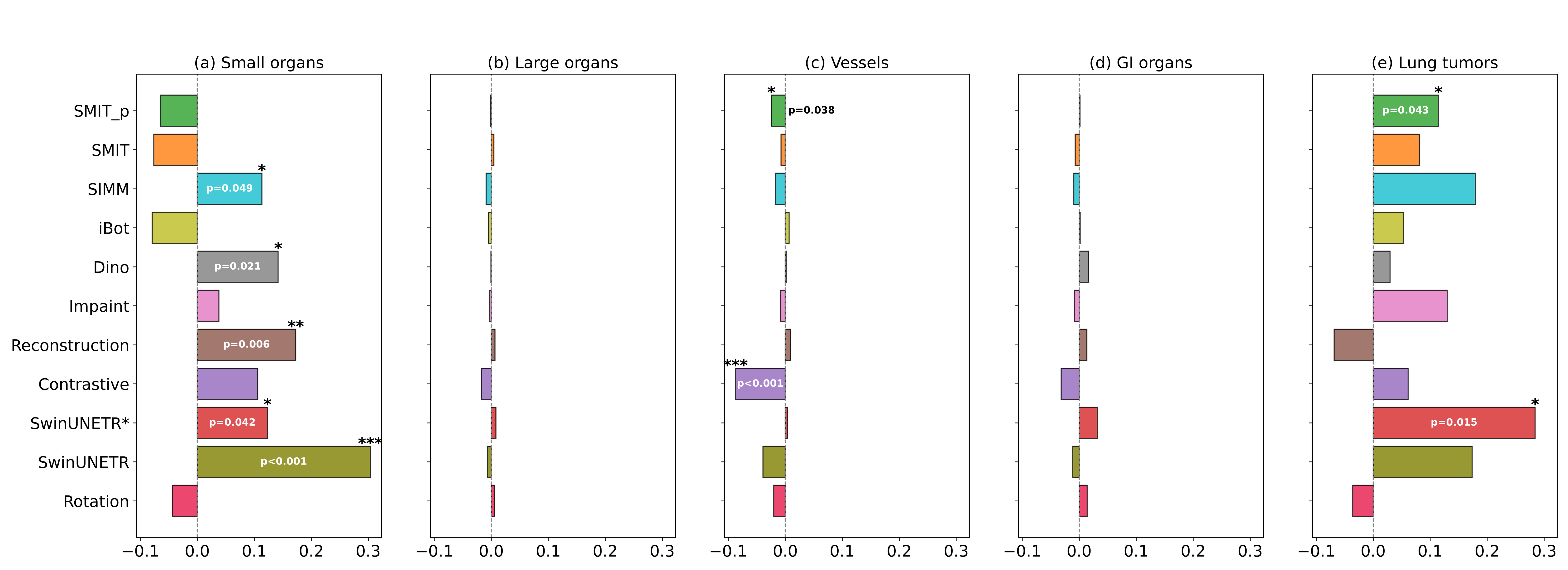}
			\caption{\small DSC accuracy difference between models for the same structures applied to MRI and CT. Significance test results are indicated as *: p $<$ 0.05; **: p $<$ 0.01; ***: p $<$ 0.001.} \label{fig:CT_MRI_Diff}
		\end{center}
	\end{figure*}

\subsection{Disaggregated analysis of performance variations across organ categories}

Performance differences of the individual models for the organ groups and lung tumor are shown for CT (Figure~\ref{fig:radar_plot}(a,b)) and MRI (Figure~\ref{fig:radar_plot}(c,d)). SMIT and SMIT$_{\mathrm{p}}$ more accurately segmented small organs than all other methods from CT and MRI. All methods performed similarly for large organs from both CT and MRI. SMIT and SimMIM were only slightly more accurate than other methods for vessels and GI organs. 
\\
Figure~\ref{fig:CT_MRI_Diff} shows the difference in accuracy between CT and MRI for the various analyzed models. Dino (p $=$ 0.021), SimMIM (p $=$ 0.049), SwinUNETR$^*$ (p $=$0.042), and SwinUNETR (p $<$0.001), as well as reconstruction (p $=$ 0.006) produced significantly more accurate segmentation of small organs from MRI compared to CT. The same models showed similar performance on CT and MRI for large organs, GI organs, and vessels. Contrastive pretraining resulted in significantly higher accuracy in CT compared to MRI (p$=$ 0.001). SMIT$_{\mathrm{p}}$ (p$=$0.038), a larger model compared to SMIT also resulted in higher accuracy on CT compared to MRI for vessels. 
\\
All SSL methods except reconstruction and rotation prediction segmented tumors more accurately from MRI than CT, despite fewer training examples used for MRI. Both SwinUNETR$^*$ (p $=$ 0.015) and SMIT$_{\mathrm{p}}$ (p $=$ 0.043) produced significantly more accurate tumor segmentation from MRI compared CT.

	\begin{figure*}[htp]
		\begin{center}
			\includegraphics[width=\textwidth]{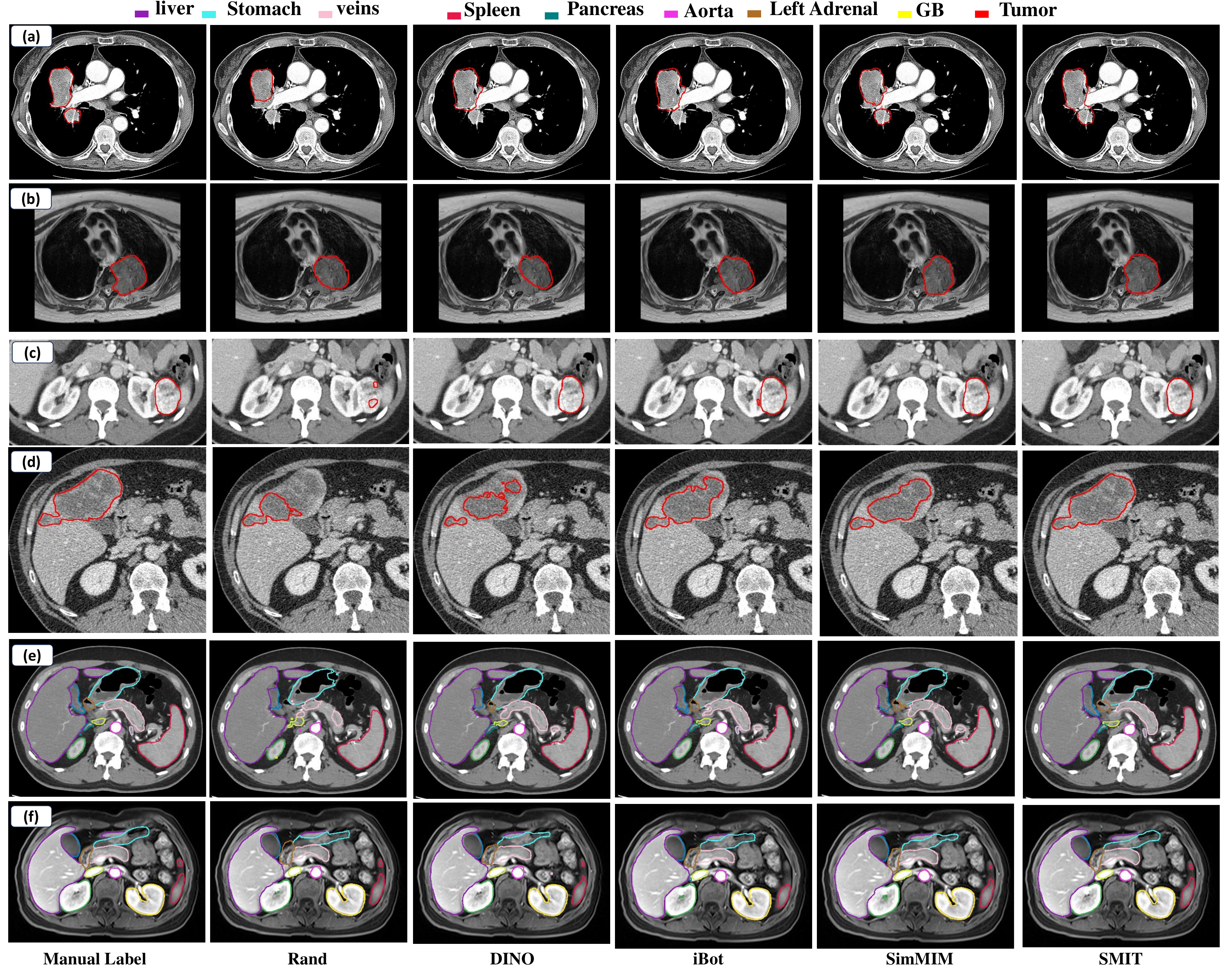}
			\caption{\small Segmentations produced by various models for representative examples:(a) lung tumor from CT, (b) lung tumor from MRI, (c) kidney tumor, (d) liver tumor, (e) abdominal organs from CT, (f) abdominal organs from MRI. } \label{fig:seg_overlay_MRI}
		\end{center}
	\end{figure*}

 	\begin{figure}[tp]
		\begin{center}
			\includegraphics[width=0.85\textwidth]{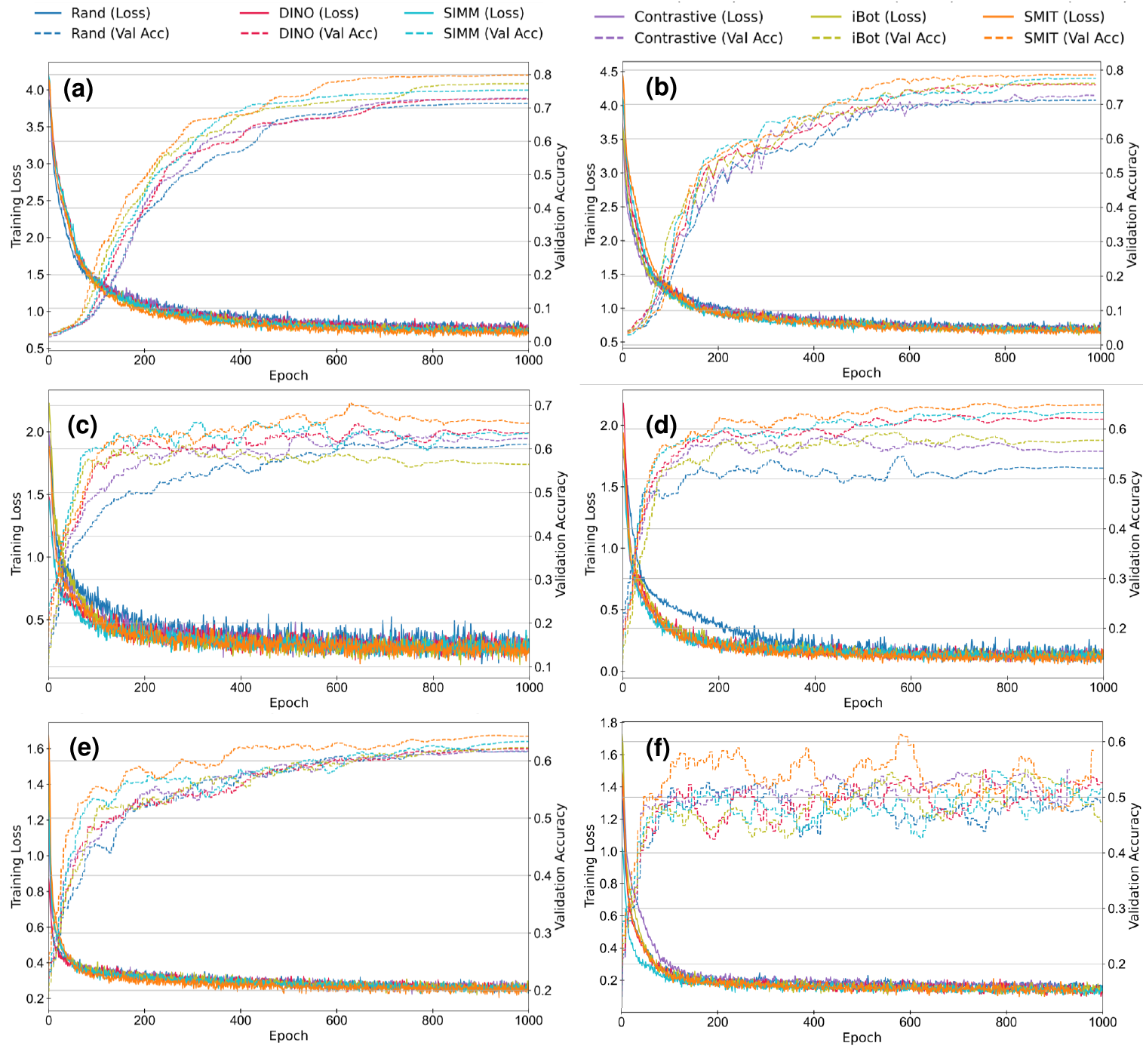}
			\vspace{-0.05cm}\setlength{\belowcaptionskip}{-0.8cm}\setlength{\abovecaptionskip}{0.08cm}\caption{\small  Training loss and validation accuracy curves for analyzed pretrained models applied to segmenting (a) abdomen organs from CT (b) abdomen organs from MRI (c) liver tumor from CT (d) kidney tumor from CT (e) lung tumor from CT, and (f) lung tumor from MRI.} \label{fig:train_curve}
		\end{center}
	\end{figure}


\subsubsection{Qualitative results}
Figure~\ref{fig:seg_overlay_MRI} shows representative cases with segmentations produced by four most accurate methods. Supervised or scratch trained model initialized with random weights is shown for baseline comparison. As shown, all pretrained models were similarly accurate when used to segment the lung tumor (Figure~\ref{fig:seg_overlay_MRI} b) as well as organs (Figure~\ref{fig:seg_overlay_MRI} f) from MRI. Pretrained models were similarly accurate for prominently appearing tumor in the kidney (Figure~\ref{fig:seg_overlay_MRI} c) and the abdominal organs from CT (Figure~\ref{fig:seg_overlay_MRI} e). On the other hand, SMIT and SimMIM more accurately segmented tumors with lower soft tissue contrast such as the lung and liver cancers from CT (Figure~\ref{fig:seg_overlay_MRI} a,d).

\subsection{Fine-tuning efficiency in downstream tasks}

Efficiency of fine-tuning the pretrained models to the downstream tasks were evaluated by assessing convergence speed and few-shot training accuracy.  

\subsubsection{Convergence speed}
Figure~\ref{fig:train_curve} shows the training curves for multiple downstream tasks applied to models pretrained with contrastive, DINO, iBOT, SimMIM, SMIT, and scratch training. Pretrained models converged faster than scratch training. SMIT converged the fastest for all tasks including lung tumor segmentation from CT and MRI, indicated by higher validation accuracy achieved at earlier epochs than other models (Figure~\ref{fig:train_curve} (e,f)). Although contrastive pretraining showed an initial fast convergence than iBOT, its accuracy saturated, ultimately resulting in less accurate performance than iBOT for liver (Figure~\ref{fig:train_curve} (c)), kidney (Figure~\ref{fig:train_curve} (d)), and lung tumors (Figure~\ref{fig:train_curve} (f)). 

\subsection{Segmentation accuracy in few-shot training regimes}
Performance gap between few- and full-shot training measured the relative difference between a model's accuracy \( acc_{m} \) compared to reference accuracy \( acc_{ref} \) corresponding to the highest accuracy achieved by any model for a given task:
\[
    \text{Performance Gap (\%)} = \frac{acc_{m} - acc_{ref}}{acc_{ref}} \times 100\%.
\]

\begin{figure}[tp]
		\begin{center}
			\includegraphics[width=0.85\textwidth]{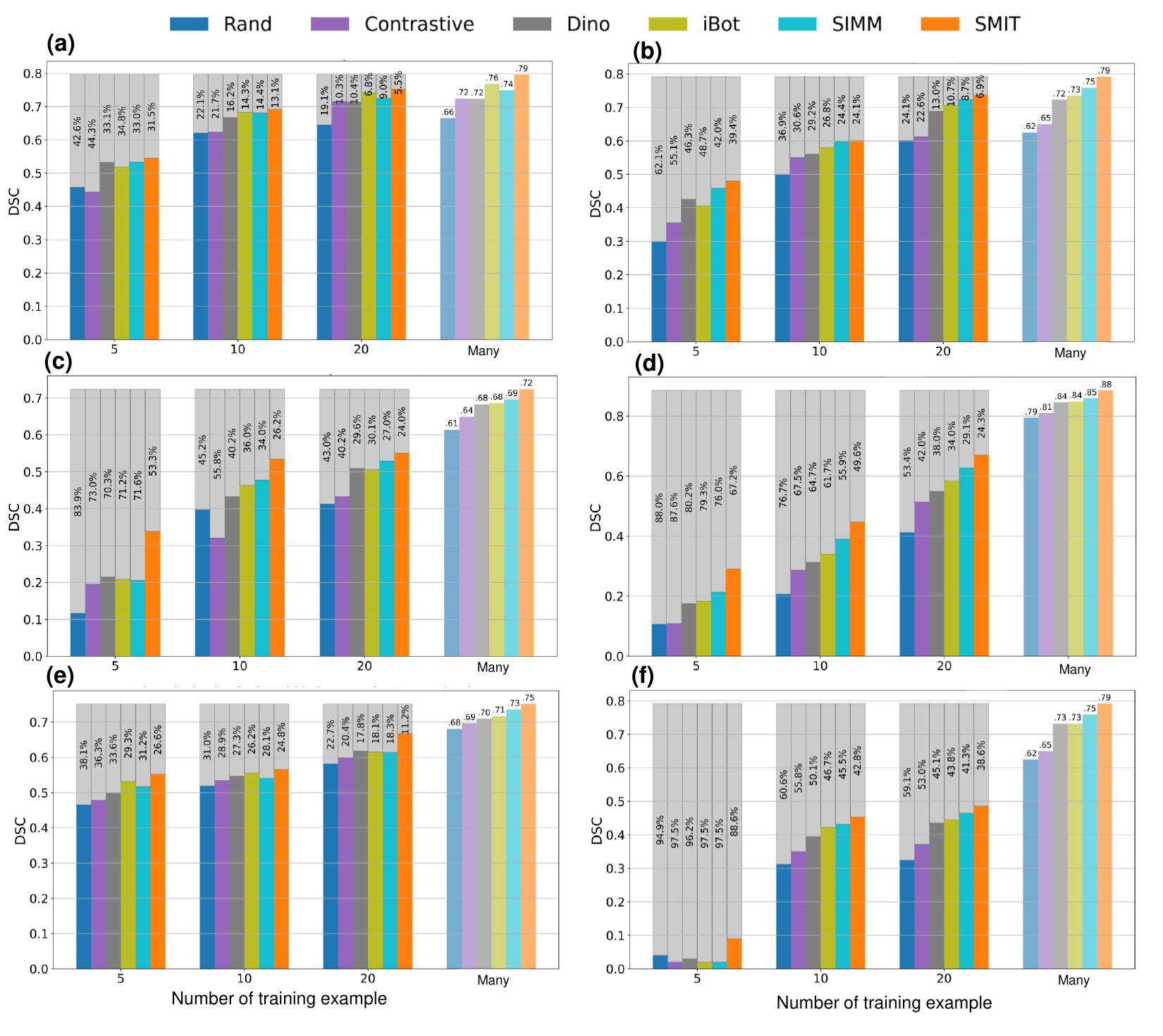}
			\vspace{-0.05cm}\setlength{\belowcaptionskip}{-0.8cm}\setlength{\abovecaptionskip}{0.08cm}\caption{\small  Few-shot and many-shot accuracies with performance gap (\%) shown for segmenting (a) abdomen organs from CT (b) abdomen organs from MRI (c) lung tumors from CT (d) kidney tumors from CT (e) liver tumors from CT, and (f) lung tumors from MRI. } \label{fig:few_shot_analysis}
		\end{center}
	\end{figure}

The few- and many-shot accuracies and the performance gaps are shown in Figure~\ref{fig:few_shot_analysis}. SMIT produced the lowest performance gap and was the most accurate in all few-shot settings for all analyzed tasks. It also showed substantial accuracy improvement over all other methods in the 20-shot training scenario, indicating capability of these models to quickly adapt features for the downstream tasks. Large differences between models were noted for lung (Figure~\ref{fig:few_shot_analysis}(c,f)) and kidney tumors (Figure~\ref{fig:few_shot_analysis}(d)), especially in the 5-shot setting. Among cancers, liver cancer segmentation resulted in the smallest performance gaps for all models. Smallest performance gaps for all models occurred for multi-organ segmentation from CT. Performance differences between models disappeared with increasing numbers of examples, indicating the specific SSL approach had limited benefit in many-shot setting with sufficient numbers of examples for organ segmentation. Larger performance gaps occurred in 5- and 10-shot cases for organ segmentation from MRI compared to CT for all models, indicating the difficulty of learning from a different modality than used in pretraining.  

\begin{figure*}[htp]
    \begin{center}
	\includegraphics[width=\textwidth]{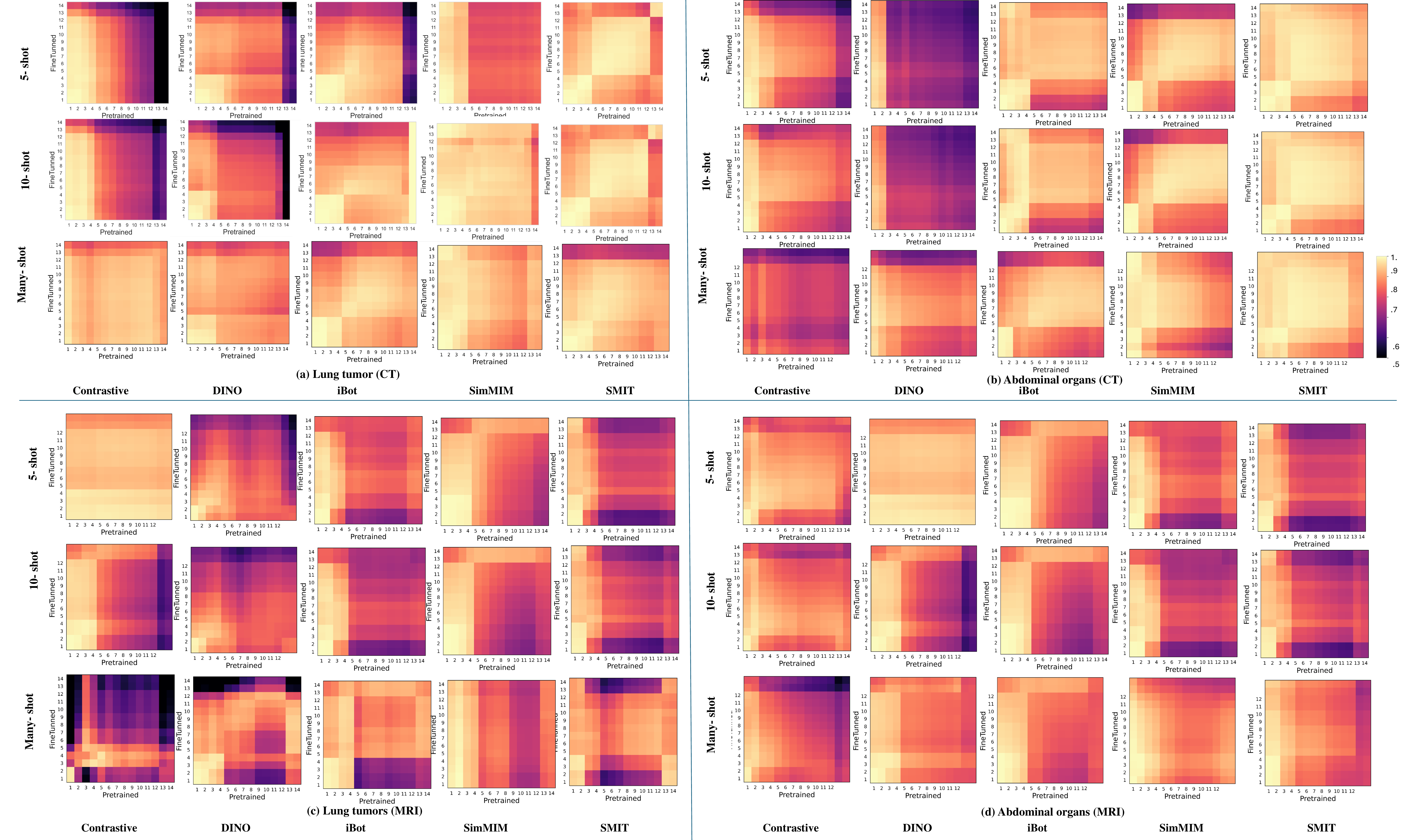}
	\vspace{-0.05cm}\setlength{\belowcaptionskip}{-0.8cm}\setlength{\abovecaptionskip}{0.08cm}\caption{\small Feature reuse analysis performed using CKA comparing pretrained versus finetuned models for 5-, 10-, and Many-shot regimes applied to segmenting (a) lung tumors from CT, (b) abdomen organs from CT, (c) lung tumors from MRI, and (d) abdomen organs from MRI. } \label{fig:CKA_analysis_Swin_full_few_shot}
    \end{center}
\end{figure*}

\subsection{Feature reuse analysis}

\subsubsection{Feature reuse for CT tasks}
SMIT pretrained models resulted in higher feature reuse across many- and few-shot (5, 10) regimes for both lung and organs segmentation from CT (Figure~\ref{fig:CKA_analysis_Swin_full_few_shot} a,b). In both cases, the pattern of feature reuse was similar with high feature reuse in low- and mid-level layers and large feature dissimilarity occurring in the higher level (13 to 14) feature layers. SimMIM showed higher feature reuse in the 10- and many-shot regime for lung tumor segmentation but showed highly variable patterns of feature reuse across the training regimes. Both contrastive and DINO methods showed very low feature reuse in the few-shot regimes for both tumor and organ segmentation tasks. iBOT resulted in higher feature reuse compared to DINO and contrastive pretraining but resulted in more feature changes across all layers compared to SimMIM and SMIT methods. 

\subsubsection{Feature reuse for MR tasks}
As opposed to CT, all models showed large feature differences between pretraining and fine tuning at all layers for both tumor and organ segmentations (Figure~\ref{fig:CKA_analysis_Swin_full_few_shot} c,d). SMIT and SimMIM resulted in similar patterns of feature reuse between 5- and 10-shot training regimes for both organs and lung tumor segmentation. However, feature reuse increased in the low (1 to 2) and mid-level layers (5-9) in the many-shot setting for both method for organs segmentation. SMIT showed higher feature reuse in the many-shot setting for lung tumor segmentation in low (1-2) and mid-level (8-12) layers. iBOT resulted in high feature reuse in the low level (1-4) layers for organs segmentation. It also similar patterns of feature reuse between few- and many-shot settings for both lung and organs segmentations. iBOT was the third most accurate method for organs and tumor segmentation from MRI in few- and many-shot settings. Contrastive method showed highly similar patterns of feature reuse for organ segmentation between few- and many-shot settings. However, it resulted in the worst convergence of all analyzed methods (Figure~\ref{fig:train_curve}) and was the least accurate method (Figure~\ref{fig:few_shot_analysis}, indicating it was less effective in extracting features applicable to different modality tasks. Contrastive method also resulted in higher feature reuse for the low-level features (layers 1-4) in 10-shot and all feature layers in the 5-shot setting for lung tumor segmentation, and a highly different feature reuse pattern in the many-shot setting. Its accuracy was substantially lower in the 5-shot compared to 20-shot setting. DINO showed large differences in patterns of feature reuse between 5-, 10- and many shot setting for both organs and tumors. 

	\begin{figure*}[tp]
		\begin{center}
			\includegraphics[width=\textwidth]{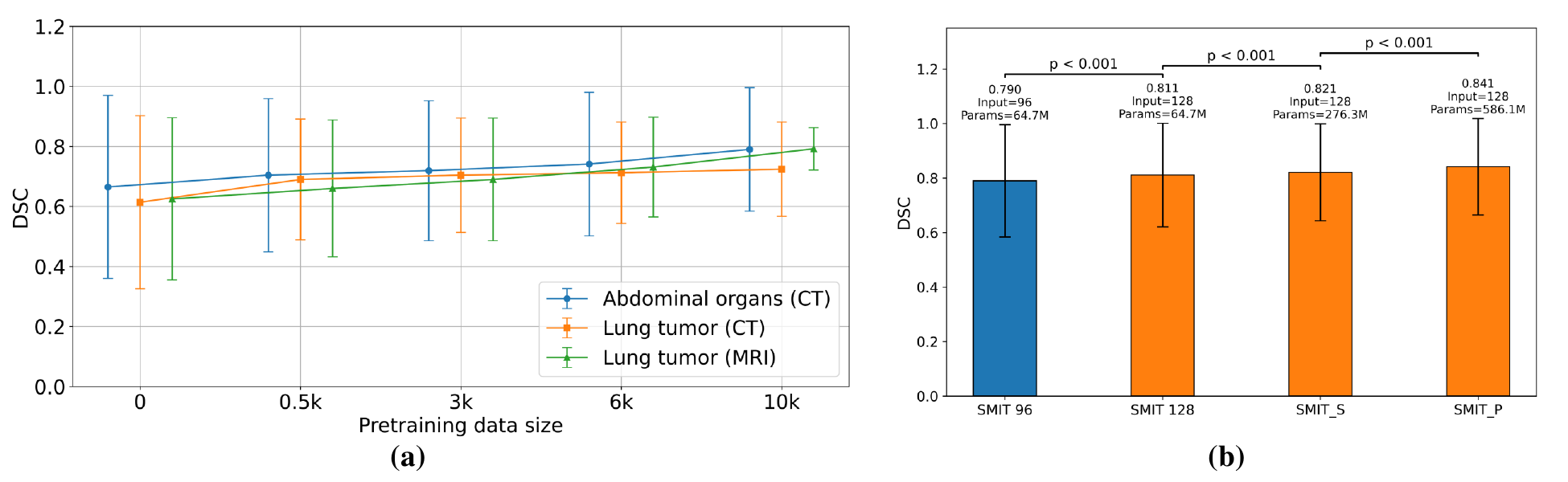}
			\vspace{-0.05cm}\setlength{\belowcaptionskip}{-0.8cm}\setlength{\abovecaptionskip}{0.08cm}\caption{\small  (a) Impact of pretraining data size on SMIT model segmentation accuracy for organs segmentation from CT, lung tumor segmentation from CT and MRI. (b) Impact of model size/capacity on multi-organ segmentation accuracy from CT. } \label{fig:PT_size}
		\end{center}
	\end{figure*}

\subsection{Design experiments} 

\subsubsection{Impact of pretraining data}
SwinUNETR and SwinUNETR$^*$ were evaluated for multiple tasks including CT and MRI organs and tumor segmentation. SwinUNETR$^*$ outperformed the published checkpoint for organs segmentation from CT and MRI as well as lung tumor segmentation from MRI, indicating pretraining with more datasets is beneficial (Figure~\ref{fig:over_all_acc} (a,b,d)). However, both models were similarly accurate for segmenting lung tumors from CT. Disaggregated analysis of accuracy differences between MRI and CT showed significant accuracy improvements for small organs with both methods, albeit SwinUNETR showed significant differences between modalities compared to SwinUNETR$^*$ (Figure~\ref{fig:CT_MRI_Diff}(a)). On the other hand, SwinUNETR$^*$ showed significant accuracy differences for lung tumor segmentation between MRI and CT (p $=$ 0.015) compared to SwinUNETR (Figure~\ref{fig:CT_MRI_Diff}(e)). 
\\
Figure~\ref{fig:PT_size} (a) shows segmentation accuracy achieved by SMIT when pretrained with varying numbers of pretraining data including 500, 3000, 6000, and 10000 cases. The model performance with scratch training (shown by pretraining size of 0) is also shown. As shown, increasing the pretraining data size improved accuracy across all three tasks. Scratch trained models were the least accurate with wide confidence intervals. Average organs segmentation accuracy increased from 0.66 for scratch trained to 0.79 for the same organs when using 10K SSL pretrained model. Similarly, accuracy for lung tumor segmentation increased from 0.61 to 0.725 for CT and 0.625 to 0.79 for MRI. Pretraining with larger number of samples also reduced the confidence intervals, thus resulting in more consistently accurate segmentations for CT and MRI tasks. 

\subsubsection{Impact of model capacity/size}

The experiment evaluated the Swin backbone of varying sizes pretrained using the default 10K pretraining data to segment abdominal organs from CT. The default Swin network (SMIT$_{96}$) used a $[2,2,8,2]$ configuration and accepts an input image of size 96$\times$96$\times$96 voxels and produced a feature embedding size of 48 and 64.7M parameters. The same Swin configuration was used to train a model accepting a larger input image of size 128$\times$128$\times$128 voxels or SMIT$_{128}$ to evaluate the impact of image size. SMIT$_{128}$ had a feature embedding size of 48 and 64.7M parameters. Two larger variants of the Swin model were created by scaling the stage 3 depth to produce SMIT$_{S}$ with a configuration $[2,2,18,2]$, feature embedding size of 96, and 276.3M parameters, and SMIT$_{\mathrm{p}}$ with a depth of [2,2,40,4], with a feature embedding size of 128, and 586.1M parameters. 
\\
SMIT$_{\mathrm{p}}$ was trained to segment organs and lung tumor from CT and MRI. It demonstrated significantly higher accuracy than SMIT for organs from CT (p$<$0.001) as well as MRI (p$<$0.001) as shown in Figure~\ref{fig:over_all_acc}(a,b). The same model was also more accurate than the SMIT model for segmenting lung tumors from CT (DSC of 0.75, p$<$0.001) and MRI (DSC of 0.81, p$<$0.001). SMIT$_{\mathrm{p}}$ also showed a significant accuracy improvement for segmenting lung tumors from MRI compared to CT (p $=$ 0.043) as shown in Figure~\ref{fig:CT_MRI_Diff}(e). The accuracy gains from SMIT were noted for every model configuration that either accepted a larger input image or had a bigger model size for multi-organ segmentation from CT (Figure~\ref{fig:PT_size} (b)). Specifically, accuracy increased from an average DSC of 0.79 to 0.811 for the same model architecture using a larger input image (p $<$ 0.001) and to 0.821 for SMIT$_{S}$. 

\begin{table*}[tp]
\centering
\scriptsize
\caption{Multi-organ and lung tumor segmentation accuracy when increasing fine tuning data size. SP: spleen, RK, LK: right, left kidney; GB: gall bladder; ESO: esophagus; LV: liver; STO: stomach; AOR: aorta; IVC: inferior vena cava; PAN: pancreas; RA, LA: right, left adrenal; Duo: Duodenum; BLD: bladder; Pros: prostate gland. }
\setlength{\tabcolsep}{1.5pt}
\renewcommand{\arraystretch}{0.65}
\begin{tabular}{lccccccccccccccc c|c}
\toprule
\textbf{Model} & {SP}& { RK  }& {  LK  }&{GB}&{ESO}&{LV}&{STO}&{AOR}&{IVC}&{PAN}&{RA}&{LA}&{Duo}&{BLD}&{Pros}&{AVG (organs)}&{Tumor} \\
\midrule
\multirow{2}{*}{\textbf{Rand}} & 0.95 & 0.93 & 0.94 & 0.75 & 0.77 & 0.96 & 0.87 & 0.91 & 0.84 & 0.80 & 0.69 & 0.68 & 0.72 & 0.81 & 0.76  & 0.82 & 0.66\\
 & \textcolor{gray}{0.03} & \textcolor{gray}{0.14} & \textcolor{gray}{0.02} & \textcolor{gray}{0.27} & \textcolor{gray}{0.10} & \textcolor{gray}{0.01} & \textcolor{gray}{0.15} & \textcolor{gray}{0.04} & \textcolor{gray}{0.05} & \textcolor{gray}{0.10} & \textcolor{gray}{0.13} & \textcolor{gray}{0.14} & \textcolor{gray}{0.15} & \textcolor{gray}{0.22} & \textcolor{gray}{0.21} &  \textcolor{gray}{0.10}& \textcolor{gray}{0.27} \\
\midrule
\multirow{2}{*}{\textbf{Contrastive}} & 0.95 & 0.92 & 0.94 & 0.79 & 0.82 & 0.97 & 0.90 & 0.94 & 0.88 & 0.84 & 0.74 & 0.74 & 0.77 & 0.86 & 0.79  & 0.85& 0.70 \\
 & \textcolor{gray}{0.06} & \textcolor{gray}{0.15} & \textcolor{gray}{0.09} & \textcolor{gray}{0.27} & \textcolor{gray}{0.10} & \textcolor{gray}{0.03} & \textcolor{gray}{0.13} & \textcolor{gray}{0.04} & \textcolor{gray}{0.04} & \textcolor{gray}{0.09} & \textcolor{gray}{0.10} & \textcolor{gray}{0.11} & \textcolor{gray}{0.14} & \textcolor{gray}{0.18} & \textcolor{gray}{0.21}  & \textcolor{gray}{0.09}& \textcolor{gray}{0.22} \\
\midrule
\multirow{2}{*}{\textbf{SwinUNETR}} & 0.94 & 0.92 & 0.94 & 0.78 & 0.81 & 0.96 & 0.90 & 0.94 & 0.88 & 0.83 & 0.73 & 0.74 & 0.76 & 0.83 & 0.78  & 0.84 & 0.68\\
 & \textcolor{gray}{0.07} & \textcolor{gray}{0.15} & \textcolor{gray}{0.08} & \textcolor{gray}{0.26} & \textcolor{gray}{0.10} & \textcolor{gray}{0.03} & \textcolor{gray}{0.12} & \textcolor{gray}{0.04} & \textcolor{gray}{0.06} & \textcolor{gray}{0.10} & \textcolor{gray}{0.11} & \textcolor{gray}{0.13} & \textcolor{gray}{0.14} & \textcolor{gray}{0.21} & \textcolor{gray}{0.22}  & \textcolor{gray}{0.09} & \textcolor{gray}{0.24}\\
\midrule
\multirow{2}{*}{\textbf{SwinUNETR$^*$}} & 0.96 & 0.93 & 0.95 & 0.79 & 0.82 & 0.97 & 0.90 & 0.94 & 0.88 & 0.84 & 0.74 & 0.75 & 0.78 & 0.85 & 0.78 & 0.85 & 0.71 \\
 & \textcolor{gray}{0.04} & \textcolor{gray}{0.14} & \textcolor{gray}{0.04} & \textcolor{gray}{0.26} & \textcolor{gray}{0.09} & \textcolor{gray}{0.01} & \textcolor{gray}{0.13} & \textcolor{gray}{0.04} & \textcolor{gray}{0.05} & \textcolor{gray}{0.10} & \textcolor{gray}{0.09} & \textcolor{gray}{0.11} & \textcolor{gray}{0.14} & \textcolor{gray}{0.19} & \textcolor{gray}{0.22} & \textcolor{gray}{0.09} & \textcolor{gray}{0.21} \\
\midrule
\multirow{2}{*}{\textbf{Inpaint}} & 0.95 & 0.93 & 0.93 & 0.78 & 0.81 & 0.97 & 0.91 & 0.94 & 0.88 & 0.84 & 0.73 & 0.74 & 0.77 & 0.86 & 0.80 & 0.70 & 0.85 \\
 & \textcolor{gray}{0.07} & \textcolor{gray}{0.14} & \textcolor{gray}{0.10} & \textcolor{gray}{0.27} & \textcolor{gray}{0.08} & \textcolor{gray}{0.03} & \textcolor{gray}{0.13} & \textcolor{gray}{0.03} & \textcolor{gray}{0.04} & \textcolor{gray}{0.10} & \textcolor{gray}{0.09} & \textcolor{gray}{0.11} & \textcolor{gray}{0.14} & \textcolor{gray}{0.16} & \textcolor{gray}{0.14} & \textcolor{gray}{0.09}& \textcolor{gray}{0.21}  \\
\midrule
\multirow{2}{*}{\textbf{Reconstruction}} & 0.96 & 0.93 & 0.95 & 0.78 & 0.79 & 0.97 & 0.90 & 0.92 & 0.87 & 0.83 & 0.73 & 0.73 & 0.76 & 0.85 & 0.77 & 0.84 & 0.71 \\
 & \textcolor{gray}{0.03} & \textcolor{gray}{0.14} & \textcolor{gray}{0.02} & \textcolor{gray}{0.26} & \textcolor{gray}{0.09} & \textcolor{gray}{0.01} & \textcolor{gray}{0.13} & \textcolor{gray}{0.04} & \textcolor{gray}{0.05} & \textcolor{gray}{0.10} & \textcolor{gray}{0.10} & \textcolor{gray}{0.11} & \textcolor{gray}{0.14} & \textcolor{gray}{0.19} & \textcolor{gray}{0.24} & \textcolor{gray}{0.09}& \textcolor{gray}{0.21}  \\
\midrule
\multirow{2}{*}{\textbf{Rotation}} & 0.95 & 0.93 & 0.94 & 0.77 & 0.79 & 0.97 & 0.90 & 0.92 & 0.86 & 0.83 & 0.73 & 0.72 & 0.75 & 0.84 & 0.77 & 0.83 & 0.69 \\
 & \textcolor{gray}{0.03} & \textcolor{gray}{0.13} & \textcolor{gray}{0.04} & \textcolor{gray}{0.26} & \textcolor{gray}{0.10} & \textcolor{gray}{0.01} & \textcolor{gray}{0.13} & \textcolor{gray}{0.04} & \textcolor{gray}{0.06} & \textcolor{gray}{0.10} & \textcolor{gray}{0.09} & \textcolor{gray}{0.12} & \textcolor{gray}{0.15} & \textcolor{gray}{0.20} & \textcolor{gray}{0.21} & \textcolor{gray}{0.09} & \textcolor{gray}{0.20} \\
\midrule
\multirow{2}{*}{\textbf{Dino}} & 0.96 & 0.94 & 0.96 & 0.79 & 0.83 & 0.97 & 0.90 & 0.94 & 0.89 & 0.85 & 0.76 & 0.76 & 0.78 & 0.88 & 0.82 & 0.86 & 0.71 \\
 & \textcolor{gray}{0.03} & \textcolor{gray}{0.14} & \textcolor{gray}{0.02} & \textcolor{gray}{0.27} & \textcolor{gray}{0.10} & \textcolor{gray}{0.01} & \textcolor{gray}{0.14} & \textcolor{gray}{0.03} & \textcolor{gray}{0.05} & \textcolor{gray}{0.09} & \textcolor{gray}{0.09} & \textcolor{gray}{0.12} & \textcolor{gray}{0.14} & \textcolor{gray}{0.14} & \textcolor{gray}{0.14} & \textcolor{gray}{0.08} & \textcolor{gray}{0.18} \\
\midrule
\multirow{2}{*}{\textbf{iBOT}} & 0.96 & 0.94 & 0.95 & 0.79 & 0.83 & 0.97 & 0.90 & 0.94 & 0.89 & 0.85 & 0.76 & 0.76 & 0.79 & 0.88 & 0.82 & 0.86 & 0.71 \\
 & \textcolor{gray}{0.04} & \textcolor{gray}{0.14} & \textcolor{gray}{0.03} & \textcolor{gray}{0.27} & \textcolor{gray}{0.10} & \textcolor{gray}{0.01} & \textcolor{gray}{0.13} & \textcolor{gray}{0.03} & \textcolor{gray}{0.05} & \textcolor{gray}{0.09} & \textcolor{gray}{0.09} & \textcolor{gray}{0.12} & \textcolor{gray}{0.14} & \textcolor{gray}{0.14} & \textcolor{gray}{0.17} & \textcolor{gray}{0.08} & \textcolor{gray}{0.17} \\
\midrule
\multirow{2}{*}{\textbf{SimMIM}} & 0.95 & 0.93 & 0.94 & 0.79 & 0.82 & 0.96 & 0.90 & 0.93 & 0.88 & 0.84 & 0.75 & 0.76 & 0.80 & 0.88 & 0.82 & 0.85 & 0.72 \\
 & \textcolor{gray}{0.03} & \textcolor{gray}{0.14} & \textcolor{gray}{0.03} & \textcolor{gray}{0.27} & \textcolor{gray}{0.10} & \textcolor{gray}{0.01} & \textcolor{gray}{0.13} & \textcolor{gray}{0.04} & \textcolor{gray}{0.04} & \textcolor{gray}{0.09} & \textcolor{gray}{0.08} & \textcolor{gray}{0.11} & \textcolor{gray}{0.10} & \textcolor{gray}{0.13} & \textcolor{gray}{0.16}& \textcolor{gray}{0.08}  & \textcolor{gray}{0.18} \\
\midrule
\multirow{2}{*}{\textbf{SMIT}} & 0.96 & 0.94 & 0.96 & \textbf{0.80} & 0.83 & 0.97 & \textbf{0.92} & 0.94 & 0.89 & \textbf{0.86} & \textbf{0.77} & \textbf{0.77} & \textbf{0.81} & \textbf{0.90} & 0.82 & \textbf{0.87}& \textbf{0.73}  \\
 & \textcolor{gray}{0.03} & \textcolor{gray}{0.14} & \textcolor{gray}{0.02} & \textcolor{gray}{0.26} & \textcolor{gray}{0.09} & \textcolor{gray}{0.01} & \textcolor{gray}{0.12} & \textcolor{gray}{0.03} & \textcolor{gray}{0.05} & \textcolor{gray}{0.09} & \textcolor{gray}{0.08} & \textcolor{gray}{0.11} & \textcolor{gray}{0.12} & \textcolor{gray}{0.12} & \textcolor{gray}{0.17} & \textcolor{gray}{0.08}  & \textcolor{gray}{0.17}\\
\midrule
\bottomrule
\end{tabular}

\label{tab:swin_more_data}
\end{table*}

\subsubsection{Does increasing fine-tuning data size reduce the impact of pretraining method?}
Finally, we evaluated whether increasing fine tuning data size to a sufficiently large number of cases reduced the accuracy gap between pretraining methods. For this purpose, the Swin pretrained models were fine tuned with a considerably large number of labeled examples consisting of 350 cases for lung tumors and 100 cases for multi-organs segmentation from CT scans. As shown in Table \ref{tab:swin_more_data}, the relative differences between the various models disappeared with all models achieving similar accuracies for organs. SMIT resulted in a slightly higher average accuracy of 0.87, compared to the least accurate pretrained model with an average accuracy of 0.85. Even the scratch trained model resulted in an average accuracy of 0.82, demonstrating that the benefit of pretraining can be overcome with larger number of examples. On the other hand, the relative advantage of pretraining was evident for tumor segmentation with all pretrained models achieving more accurate segmentation than scratch trained model. The published checkpoint SwinUNETR that was pretrained with half the number of cases as used for all other models including SwinUNETR$^{*}$ was the least accurate pretrained method.

\section{Discussion}

In this work, we benchmarked a number of commonly used SSL pretraining methods within a Swin Transformer encoder--convolutional decoder pipeline (SwinUNETR-style) for organs and tumor segmentation from CT and MRI. Our analysis showed that pretraining methods using MIM such as SMIT and SimMIM resulted in more accurate segmentation in few- and many-shot settings across all tasks even in tasks involving modality such as MRI differing from the pretraining modality. SMIT, which employs multi-objective pretraining was the most accurate method, indicating the benefit of multi-objective pretraining. On the other hand, methods that primarily focused on global feature representations were the least accurate. Complementing previous SSL benchmarks that primarily focused on within-modality evaluation, our work provides a systematic comparison of feature reuse, training efficiency, and cross-modality transfer in both data-limited and many-shot regimes for segmentation tasks of varying complexity.  
\\
Prior works have considered the problem of feature reuse and transferability to understand model performance in domains that diverge from the pretraining domain~\citep{Matsoukas2022,Goyal2021,Haghighi2024_MedIA}. Most works studied the problem of transferability to medical imaging domains following supervised pretraining with photographic images~\citep{Matsoukas2022,Goyal2021} and showed that increasing distance of the target domain from the source or pretraining domain reduces transferability. Supervised pretrained models like the MedSAM3D~\citep{wang2024sam}, UniverSeg~\citep{Butoi2023} have also shown to be effective for generating segmentations on diverse imaging modalities, albeit such models require either bounding box/point-click prompts or support sets at test time. 
\\
The works most closely related to our work are the methods using SSL pretraining with medical images~\citep{Haghighi2024_MedIA,jiang2022self,Jiang_ISBI_2025} that showed transferability of models pretrained with CT datasets to non-CT imaging modalities. Our work enhances the aforementioned works by studying why such transferability occurs for both organs and tumors occurring in different anatomic locations. Our results show that MIM-based tasks are highly effective in extracting locality specific attention information, which when complemented by tasks that extract global information leads to a richer and diverse feature representation applicable to tasks in different modalities, shown by spatial diversity of multi-head attention with SMIT. Our findings align with the finding from Haghighi et al.~\citep{Haghighi2024_MedIA} that SSL combining multiple pretext tasks achieve better downstream performance over different tasks. Results from SwinUNETR further confirm our findings that tasks primarily focused on extracting global representation, even when used in multi-objective setting may not lead to higher accuracy in downstream tasks. 
\\
As limitation, our work primarily focused on segmentation tasks and considered only CT and anatomic MRI. Extension to functional images like diffusion weighted MRI as well as positron emission tomography images is future work. Nevertheless, by training all SSL methods on the same large CT dataset and evaluating them on the same public benchmarks, this work provides a controlled, side-by-side comparison of behavior that complements prior medical SSL benchmarks. 

\section{Conclusion}
This work performed a comprehensive analysis of multiple SSL tasks to pretrain Swin based models applied to semantic segmentation of medical images for organs and tumors segmentation from CT and MRI. Our analysis showed SMIT pretraining that combines masked image modeling with token self-distillation balanced the extraction of local and global image representations, which enhanced its transferability to various tasks in data-limited and many-shot regimes including for the more challenging tumor segmentation from different modality than the pretraining modality.

\acks{Acknowledgment}
This research was partially supported by the NCI R01CA258821 and the Memorial Sloan Kettering Cancer Center Support Grant/Core Grant NCI P30CA008748 

%
\ethics{The work follows appropriate ethical standards in conducting research and writing the manuscript, following all applicable laws and regulations regarding treatment of animals or human subjects.}

\coi{We declare we don't have conflicts of interest.}

\data{The AMOS, LITS, KiTS, and Lung TCIA datasets are publicly available. The internal CT and MRI lung datasets, however, cannot be shared due to institutional policy. }

\bibliography{mybibliography}

\end{document}